\documentclass[10pt,twocolumn,letterpaper]{article}

\usepackage{cvpr}              
\definecolor{cvprblue}{rgb}{0.21,0.49,0.74}
\usepackage[pagebackref,breaklinks,colorlinks,allcolors=cvprblue]{hyperref}
\usepackage{tabularx, multirow, booktabs, array}
\newcolumntype{C}{>{\centering\arraybackslash}X}

\usepackage{multirow}
\usepackage{makecell}
\usepackage{tabularx}
\usepackage{booktabs}
\usepackage{placeins}

\def\paperID{} 
\def\confName{CVPR}
\def\confYear{2026}

\title{Learning Where to Look and How to Judge: Resolution-agnostic Image Quality Assessment with Quality-aware Saliency}

\author{
Hakan Emre Gedik$^{1}$, Shashank Gupta$^{1}$, Alan Bovik$^{1,2}$ \\
$^{1}$ The University of Texas at Austin, $^{2}$ University of Colorado Boulder \\
{\tt\small
\{hakan.gedik, shashank.gupta\}@utexas.edu \;
Alan.Bovik@colorado.edu
}
}

\begin{document}
\maketitle
\begin{abstract}
    No-reference image quality assessment (NR IQA) has recently benefited from deep and multimodal models, yet many SOTA systems still violate at least one basic requirement: they either discard critical quality cues via aggressive resizing, fail to generalize across resolutions, cannot be jointly trained on heterogeneous IQA datasets with mismatched MOS scales, or require prohibitive computation. We present \textbf{ReLIQS}, a model for \textbf{Re}solution-agnostic \textbf{L}earning for \textbf{I}mage \textbf{Q}uality with \textbf{S}aliency, which is resolution-agnostic, preserves original-resolution quality cues, learns from multiple subjective studies, and remains computationally efficient and budget-adaptive. ReLIQS is a CLIP-based multiscale patch-driven architecture that learns both \emph{where to look} and \emph{how to judge} quality. Fixed-size patches are sampled across multiple resolutions, including the original resolution, and encoded with a CLIP vision backbone. A lightweight Perceptual Importance Estimator then predicts IQA-specific importance maps to select a small set of informative patches, and a Latent Quality Axis Module aggregates their embeddings into a single image-level score. Across authentic, synthetic, and AIGC benchmarks spanning diverse resolutions and distortions, ReLIQS generalizes better than strong CNN-, CLIP-, and MLLM-based baselines with matching or reduced computational cost.
\end{abstract}    
\section{Introduction}
\label{sec:introduction}
Image quality assessment (IQA) is a long-standing yet unresolved problem. Early no-reference methods \cite{niqe, ilniqe, brisque, diivine} attempted to explicitly model the Human Visual System (HVS), but struggle to generalize to complex, commingled real-world distortions. Deep learning-based methods have enabled major steps forward by learning data-driven surrogates of the HVS, but they are fundamentally constrained by the scarcity of labeled data, since subjective studies are costly and cumbersome. As a result, modern NR IQA models almost always rely on transfer learning and large-scale pretraining \cite{musiq, liqe, unique, qpt, reiqa}. This creates a basic tension with resolution: when input resolutions drift from those seen in the pretraining regime, convolutional, transformer-based, and hybrid backbones often suffer performance drops, and operating on ultra-high-resolution inputs can be both inefficient and ineffective for these architectures. In practice, most deep IQA pipelines therefore resize images to resolutions comparable to those used during pretraining, which stabilizes performance but discards low-level information that matters for quality.

IQA is sensitive to low-level aspects such as sharpness, noise, and texture integrity, as well as high-level aspects such as semantics, composition, and color balance. Downscaling suppresses fine-grained artifacts but can strengthen a backbone’s sensitivity to global structure, since the resulting resolutions are closer to those seen during pretraining. In practice, this interplay between low-level information loss and high-level robustness suggests an “optimal” resizing scale rather than a monotonic trend. For example, in the KonIQ-10K baseline study with KonCept512 model \cite{koniq}, even for mid-resolution $768 \times 1024$ images, resizing the short image dimension to $384$ can lead to better IQA performance than using the original resolution, whereas further resizing to $224$ degrades performance.     

The information loss introduced by resizing may be tolerable for mid-resolution images. For high- and ultra-high-resolution contents, however, aggressive downsampling can remove the fine-grained cues that human observers rely on, making trained models effectively insensitive to subtle perceptual defects. This motivates our first two requirements for practical, real-world directed, deep learning-based IQA: \textbf{(i)} it should be able to perform well on arbitrary image resolutions, and \textbf{(ii)} it should preserve original resolution quality cues.  

We address these requirements with a multiscale, patch-based pipeline. In brief, we sample fixed-size patches from the original image and its resized variants, encode them with a CLIP-based vision backbone, and fuse the resulting embeddings into a single quality score. By choosing patch sizes close to the CLIP pretraining resolution, we avoid degradations that occur when feeding off-distribution resolutions into the backbone, satisfying \textbf{(i)}. At the same time, patches drawn from downsampled views capture high-level aspects of quality, while patches drawn from the original image preserve low-level details, ensuring that \textbf{(ii)} is also met.

Preserving low-level quality aspects is desirable, but enforcing exhaustive spatial coverage becomes prohibitively expensive on high- and ultra-high-resolution images, since the number of patches grows quadratically with resolution. This motivates our third requirement: \textbf{(iii)} the computational cost of the model should remain reasonable even for ultra-high-resolution images, and be tightly controllable according to the available budget.

We address this requirement by hypothesizing that many patches carry redundant quality information, so accurate assessment does not require exhaustive spatial coverage. To exploit this, we introduce a lightweight auxiliary module, trained jointly with our model, that predicts relative patch weights at each scale. These weights are used to aggregate patch embeddings into scale-level representations and to rank patches by importance. At test time, we control computational cost by evaluating only the top-$k$ most important patches per scale, where $k$ is chosen according to the available budget, thereby satisfying \textbf{(iii)}. Experiments verify that sparse, importance-guided sampling preserves accuracy while reducing computation by up to 90\% compared to exhaustive coverage in ultra-high-resolution IQA.  

Beyond preserving low-level quality cues and being computationally efficient, an IQA model should also generalize across a wide range of distortions and content. Because subjective studies are cumbersome and costly, individual IQA datasets are relatively small and each reflects its own MOS scale. Naively merging these datasets to obtain more supervision often hurts generalization, since their perceptual scales are not directly compatible. In practice, models trained with standard $\ell_1$ or $\ell_2$ losses \cite{musiq,hyperiqa,nima,maniqa,dbcnn,clipiqa} are tied to a single IQA dataset, hence a single subjective study and its specific content–distortion distribution. This motivates our final requirement: \textbf{(iv)} the model should be trainable on multiple IQA datasets. We satisfy this requirement with a training objective based on within-dataset ranking and correlation information. While MOS values are difficult to pool across datasets, within-dataset ranking and correlation terms provide reliable supervision. We experimentally validated this by showing that our model generalizes across multiple datasets with distinct content and distortion types.

Although requirements \textbf{(i)}–\textbf{(iv)} are natural goals for an off-the-shelf NR IQA model, current SOTA methods typically fail to meet at least one of them. Recently proposed MLLM-based approaches \cite{qalign, compare2score, deqa} support multi-dataset training, but rely on aggressive resizing and heavy backbones, violating \textbf{(ii)} and \textbf{(iii)}. MUSIQ \cite{musiq} includes original-resolution processing, but its transformer backbone has quadratic complexity in the input resolution, making ultra-high-resolution inference impractical and failing \textbf{(i)} and \textbf{(iii)}. Other CNN-based or hybrid models \cite{hyperiqa, nima, paq2piq, dbcnn, metaiqa, reiqa, tres, topiq, contrique} are efficient and, at their target resolutions, avoid severe information loss, yet their performance degrades when the test resolution deviates significantly from the pretraining regime, failing to satisfy \textbf{(i)}.     

Here, we explain an IQA pipeline and model that satisfies all of requirements \textbf{(i)}–\textbf{(iv)}. Beyond meeting these properties, our method achieves superior performance on NR IQA benchmarks compared to SOTA approaches, often with lower or comparable computational cost. We also introduce the notion of \textbf{IQA-specific saliency} and a training methodology to realize it, which plays a central role in learning where to look and in strictly controlling test-time computation. We hope this framework can serve as a reference point for future NR IQA models designed to meet these practical requirements.
\section{Related Work}
\label{sec:related_work}
\textbf{Image Quality Assessment.} IQA methods are commonly grouped into full-reference (FR) \cite{ssim, msssim, vif}, reduced-reference (RR) \cite{rred, reduced_ref}, and no-reference (NR) \cite{niqe, ilniqe, brisque, diivine} approaches, depending on the availability of a pristine reference. Since references are rarely present in real-world applications, NR methods are often the most practical.

\noindent\textbf{Deep Learning-based Approaches.} Deep learning has advanced IQA by enabling data-driven modeling of the HVS. Early methods trained CNNs directly on MOS \cite{cnniqa}, whereas modern CNN-based \cite{hyperiqa, nima, paq2piq, dbcnn, metaiqa}, transformer-based \cite{maniqa, musiq, liqe}, and hybrid \cite{tres} models typically rely on transfer learning with a pretraining stage that both supplies robust visual features and mitigates overfitting to scarce IQA labels. Many early deep IQA models were built on ImageNet classification pretraining \cite{imagenet}, while recent work has explored vision–language contrastive \cite{clipiqa, liqe} and quality-aware pretraining \cite{contrique, qpt, reiqa}. In this work, we adopt the OpenAI CLIP model to utilize large-scale vision–language contrastive pretraining. Unlike many recent vision backbones \cite{dino, dinov2, dinov3, moco, mocov2}, it does not rely on heavy distortion-style augmentations, which helps preserve sensitivity to image degradations.

\noindent\textbf{Multi-modal Large Language Models (MLLMs).} MLLM-based IQA has recently become popular. Beyond predicting a single quality score, these models can also produce natural-language quality explanations \cite{depict-qa}, closer to how humans describe perceived quality. To evaluate and improve such abilities, Q-Bench \cite{q-bench, q-bench2} and Q-Instruct \cite{q-instruct} introduce image–text pairs with quality-focused questions, feedback, and descriptions, going beyond traditional IQA datasets that primarily provide scalar MOS labels. 

MLLM-based methods can also be repurposed to output single quality scores similar to purely visual IQA methods. Q-Align \cite{qalign} performs softmax over predefined quality descriptors to obtain a continuous score. Compare2Score \cite{compare2score} introduced a comparison-based loss for better generalization. DeQA \cite{deqa} introduced a scheme to convert continuous MOS ground-truths to soft distributions.

Despite their ability to produce natural-language quality explanations and strong results on standard benchmarks, MLLM-based IQA models are compute-intensive and often impractical for real-world deployment. They also typically resize inputs to the fixed pretraining resolution of their vision encoder, which is especially problematic for high-resolution IQA. Our method addresses both issues while matching or surpassing MLLM-based baselines.    

\noindent\textbf{Saliency in IQA.} Saliency has been used in IQA mainly to guide models with perceptually important regions, with the goal of improving both performance and interpretability \cite{sci, eye_tracking, visual_importance, jnd_saliency, vsi, saliency_review}. However, while being statistically significant, the added value of visual saliency to IQA has been shown to be marginal \cite{eye_tracking}. Furthermore, visual saliency and IQA saliency are not necessarily identical \cite{eye_tracking}. In our framework, IQA-specific saliency is learned in a data-driven manner and used primarily to reduce compute by selecting a small set of informative patches, a role that has not been explored in prior work. In line with previous studies, it also brings a modest performance gain.

\noindent\textbf{High-Resolution IQA.} Most publicly available IQA datasets \cite{koniq, spaq, paq2piq, clive, kadid, csiq, live, agiqa-3k} contain low- or mid-resolution images, making it difficult to assess how well SOTA models handle truly high-resolution content. Since deep IQA models typically resize inputs to a fixed, relatively low resolution, fine-grained quality cues are often lost, but this limitation is partly masked on conventional benchmarks. To explicitly probe high-resolution behavior, we include the UHD \cite{aim_uhd} dataset, which contains ultra-high-resolution images with subtle perceptual defects. As shown in \cref{sec:experiments}, existing SOTA IQA methods, including recent MLLM-based approaches, exhibit a noticeable performance drop on UHD, while our model achieves the strongest performance among the compared methods.      
\section{Method} 
\label{sec:method}
\begin{figure*}[t]
  \centering
  \includegraphics[width=0.85\linewidth]{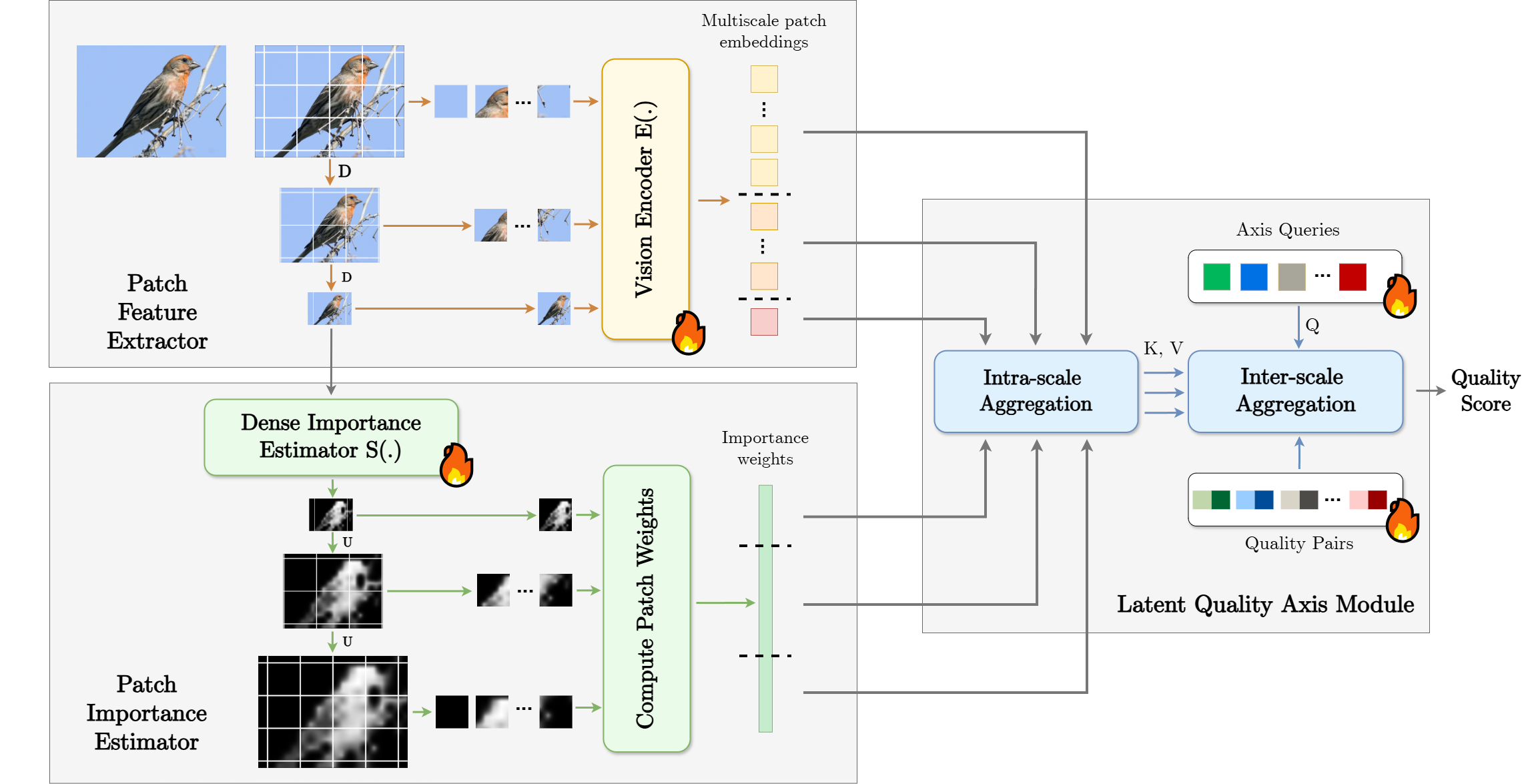}
  \caption{
    ReLIQS pipeline: patches from three scales (including original resolution). A lightweight $S(\cdot)$ on the lowest-resolution variant produces a perceptual importance map. Patches are independently encoded by $E(\cdot)$ and aggregated by LQAM.
  }
  \label{fig:diagram}
\end{figure*}

\subsection{Pipeline} 
We propose ReLIQS, a patch-based IQA framework that comprises a CLIP-based patch encoder, 
a \textbf{Perceptual Importance Estimator (PIE)} that produces a dense importance field, with which it generates patch weights, and a \textbf{Latent Quality Axis Module (LQAM)} that discovers latent quality axes and adaptively fuses multiscale information into a unified quality score. 
As in \cref{fig:diagram}, our pipeline includes:
\begin{enumerate}
    \item Sampling fixed-size patches from the original input image and its resized variants (multiscale sampling).
    \item Encoding each patch independently using a CLIP visual encoder.
    \item Computing a representation for each scale through within-scale pooling (LQAM) guided by the learned importance field (PIE).
    \item Inferring a distribution over latent quality axes for each scale using learned axis queries (LQAM).
    \item Producing the final quality prediction by adaptively combining the axis-aware predictions across scales (LQAM).
\end{enumerate}

Concretely, given an input image $\mathbf{x}^{(0)} \in \mathbb{R}^{3 \times H_0 \times W_0}$, we generate a set of its resized variants $\{\mathbf{x}^{(r)}\}_{r=1}^{R}$ while maintaining the original aspect ratio. We then sample fixed-size patches from each scale using a predefined multiscale sampling strategy, forming the patch set  $\mathbf{P} = \{ \{ \mathbf{x}_{p}^{(r)} \}_{p=1}^{c_r}  \}_{r=0}^{R}$, where $c_r$ is the number of patches sampled at scale $r$.

After constructing the patch set, the patches are independently encoded with a CLIP visual encoder $E(\cdot)$. 
\begin{equation}
    \label{eq:encoding} 
    \mathbf{e}_{p}^{(r)} = E(\mathbf{x}_{p}^{(r)}), 
\end{equation}
where $\mathbf{e}_{p}^{(r)} \in \mathbb{R}^{D}$ and $D$ is the embedding dimension. The PIE module predicts a dense importance field using a lightweight network $S(\cdot)$ applied to the smallest resized version of the image:
\begin{equation}
\label{eq:pie-small}
\mathbf{s}^{(R)} = S(\mathbf{x}^{(R)}), 
\end{equation}
where $\mathbf{s}^{(R)} \in \mathbb{R}^{H_R \times W_R}$.
The importance fields at other scales are obtained via bilinear upsampling:
\begin{equation}
\label{eq:pie-upsample}
\mathbf{s}^{(r)} = \text{Upsample}_{H_r,W_r}(\mathbf{s}^{(R)}), 
\quad r = 0,\dots,R-1.
\end{equation}

We extract importance-field patches aligned with the image patches:
\begin{equation}
\label{eq:saliency_patches}
   \omega_{p}^{(r)} = \mathbf{s}^{(r)} \big|_{\mathcal{R}_{p}^{(r)}}, 
\end{equation} 
where $\mathcal{R}_{p}^{(r)}$ denotes the spatial region of patch $p$ at scale $r$.

The patch importance weight is computed by summing the pixel-wise importance values within the patch and normalizing within each scale:
\begin{equation}
\label{eq:patch_weights}
\Omega_{p}^{(r)} = \sum_{(i,j)\in\mathcal{R}_{p}^{(r)}} \omega_{p}^{(r)}(i,j), 
\qquad w_{p}^{(r)} = \frac{\Omega_{p}^{(r)}}{\sum_{p=1}^{c_r} \Omega_p^{(r)}}.
\end{equation}

Overall, $w_{p}^{(r)}$ reflects the relative importance of patch $p$ among all patches at scale $r$. 

These relative importance weights are used for within-scale pooling, yielding a single representation per scale:
\begin{equation}
\label{eq:within-scale_aggregation}
\textbf{e}^{(r)} = \sum_{p=1}^{c_r} w_{p}^{(r)} \mathbf{e}_{p}^{(r)}.
\end{equation}

Depending on the sampling scale, patches capture complementary cues of image quality. Fine-scale patches emphasize low-level distortions such as blur, noise, and texture integrity, while coarse-scale patches convey higher-level factors such as semantic content, composition, and color balance.

Instead of relying on handcrafted quality attributes, we introduce LQAM, which learns $A$ latent quality axes and predicts how each scale contributes to them. Specifically, LQAM maintains a set of learnable quality direction pairs \(\mathbf{U}_A = \{\,(\mathbf{u}_{a-}, \mathbf{u}_{a+}) \mid \mathbf{u}_{a-}, \mathbf{u}_{a+} \in \mathbb{R}^{D}\,\}_{a=1}^{A}\), where each direction pair is associated with a latent quality axis along which the quality is assessed. Before evaluating scale representations on \(\mathbf{U}_A\),  
we project them into the axis space using learnable key and value projections:
\begin{equation}
    \label{eq:pool_kqv} 
    \mathbf{k}_{a}^{(r)} = \mathbf{K}_{a} \mathbf{e}^{(r)}, 
    \quad 
    \mathbf{v}_{a}^{(r)} = \mathbf{V}_{a} \mathbf{e}^{(r)},
\end{equation}
where \(a=1,\dots,A\) and \(\mathbf{K}_a, \mathbf{V}_a\) are linear transformations.  
The affinity of each scale representation \(\mathbf{e}^{(r)}\) to the latent axes is computed using a learned set of axis queries  
\(\mathbf{Q} = \{\mathbf{q}_a \in \mathbb{R}^{D}\}_{a=1}^{A}\):
\begin{equation}
\label{eq:pool_projection}
\beta_{a}^{(r)} =
\frac{
    \exp\!\Big(
        \operatorname{sim}\big(\mathbf{q}_{a}, \mathbf{k}_{a}^{(r)}\big) / T_{pa} 
    \Big)
}{
    \sum_{a'=1}^{A}
    \exp\!\Big(
        \operatorname{sim}\big(\mathbf{q}_{a'}, \mathbf{k}_{a'}^{(r)}\big) / T_{pa'}
    \Big)
},
\end{equation}
where \(\operatorname{sim}(\cdot,\cdot)\) denotes cosine similarity and  
\(\{T_{pa}\}_{a=1}^{A}\) are learnable temperature parameters.  
Here, \(\beta_{a}^{(r)}\) represents the probability that scale \(r\) contributes to axis \(a\).

Axis-specific quality scores are then obtained using the value projections \(\mathbf{v}_a^{(r)}\):
\begin{equation}
\label{eq:pool_latent_qa}
l_a^{(r)} =
\frac{
    \exp\!\Big(\operatorname{sim}\!\big(\mathbf{u}_{a+}, \mathbf{v}_a^{(r)}\big) / T_{qa}\Big)
}{
    \sum_{\sigma \in \{+,-\}}
    \exp\!\Big(\operatorname{sim}\!\big(\mathbf{u}_{a\sigma}, \mathbf{v}_a^{(r)}\big) / T_{qa}\Big)
}, 
\end{equation}
where  \(\{T_{qa}\}_{a=1}^{A}\) are learnable temperature parameters.  
\cref{eq:pool_latent_qa} effectively applies a softmax over each quality direction pair  
in \(\mathbf{U}_A\), assigning a score \(l_a^{(r)} \in [0,1]\) for axis \(a\) at scale \(r\).

Finally, LQAM aggregates all axis scores into a unified image quality prediction:
\begin{equation}
    \label{eq:pool_final} 
    l = \sum_{a=1}^{A} \gamma_a \sum_{r=0}^{R} \beta_{a}^{(r)}\,  l_a^{(r)},
\end{equation}
where \(\{\gamma_a\}_{a=1}^{A}\) are learnable global axis weights normalized to sum to one.

Our pipeline satisfies several key properties.

\noindent\textbf{Arbitrary resolutions and preserving low-level quality cues (i) and (ii).}
By operating on fixed-size patches, our pipeline sidesteps performance degradations that occur when input resolutions deviate from the backbone’s pretraining regime, making it effectively resolution-agnostic \textbf{(i)}. Moreover, by sampling patches not only from resized variants but also from the original-resolution image, the model retains access to all low-level quality cues, satisfying the preservation requirement \textbf{(ii)}.

\noindent\textbf{Budget-controlled scalability (iii).}  
The framework can process any number of patches at any scale, making its computational cost directly adjustable to the available budget. The lightweight PIE module (\cref{eq:pie-small,eq:pie-upsample}) efficiently estimates IQA-specific saliency maps, from which the patch weights (\cref{eq:patch_weights}) identify the most informative regions for quality assessment. We can thus allocate encoding computation to only the most perceptually relevant patches, achieving strong performance even in constrained budgets.

\noindent\textbf{Permutation invariance.}  
\cref{eq:within-scale_aggregation,eq:pool_final} are permutation-invariant, while  
\cref{eq:encoding,eq:pie-small,eq:pie-upsample,eq:saliency_patches,eq:patch_weights,eq:pool_kqv,eq:pool_projection,eq:pool_latent_qa} are permutation-equivariant. Consequently, the output remains unchanged regardless of the sampling or processing order of patches and scales, ensuring that the predicted quality depends solely on image content rather than processing order.

\subsection{Training Objective} 
To jointly train on multiple IQA datasets, we combine a margin-ranking loss and a PLCC loss in our training objective. For the dataset $d \in \{1,\dots,D\}$, given ground-truth MOS $\{g_i\}_{i=1}^{N}$ and corresponding model predictions $\{p_i\}_{i=1}^{N}$ within a mini-batch, the per-dataset margin-ranking loss term is defined as:

\begin{equation}
    \label{eq:margin-ranking} 
    \mathcal{L}_{\text{MR}}^{(d)} =
        \frac{2}{N(N-1)}\sum_{i<j} \max \Bigl(
            0,\;
            \delta -\,\operatorname{sign}\bigl(g_i - g_j\bigr)
            \bigl(p_i - p_j\bigr)
        \Bigr) 
\end{equation} 
where $\delta$ is the margin hyperparameter. For the same mini-batch, the per-dataset PLCC loss term is defined as:   
\begin{equation}
\label{eq:plcc}
\mathcal{L}_{\mathrm{PLCC}}^{(d)}
=
1 -
\frac{
    \sum_{i=1}^{N} (g_i - \bar{g})(p_i - \bar{p})
}{
    \sqrt{
        \sum_{i=1}^{N} (g_i - \bar{g})^2
        \sum_{i=1}^{N} (p_i - \bar{p})^2
    }
},
\end{equation}
where $\bar{g}$ and $\bar{p}$ denote the batch means of the ground-truth MOS and predictions, respectively. 

Both $\mathcal{L}_{\text{MR}}^{(d)}$ and $\mathcal{L}_{\text{PLCC}}^{(d)}$ are computed independently for each dataset $d \in \{1,\dots,D\}$ and then averaged:

\begin{equation}
    \label{eq:loss_average}
    \mathcal{L}_{\text{MR}} = \frac{1}{D}\sum_{d=1}^{D} \mathcal{L}_{\text{MR}}^{(d)}, 
    \qquad
    \mathcal{L}_{\text{PLCC}} = \frac{1}{D}\sum_{d=1}^{D} \mathcal{L}_{\text{PLCC}}^{(d)}.
\end{equation}
The final training objective (\cref{eq:final_loss}) combines these averaged terms using the uncertainty-based weighting strategy of \cite{uncertainty}, which adaptively balances their contributions without manual tuning:
\begin{equation}
    \label{eq:final_loss}
    \mathcal{L} =
    \frac{1}{2\sigma_{1}^{2}} \, \mathcal{L}_{\text{MR}} +
    \frac{1}{2\sigma_{2}^{2}} \, \mathcal{L}_{\text{PLCC}} +
    \log\sigma_{1} + \log\sigma_{2}, 
\end{equation}
where $\sigma_1$ and $\sigma_2$ are learnable parameters. Overall, this objective utilizes intra-dataset ranking and correlation information that remain reliable across datasets with differing perceptual scales, thereby satisfying our final requirement \textbf{(iv)} by enabling joint training on multiple IQA datasets. Loss ablations are provided in the Supplementary Material.

\section{Experiments}
\label{sec:experiments} 
\begin{table*}[tb]
    \centering
    \renewcommand{\arraystretch}{0.7}
    \setlength{\tabcolsep}{2.25pt}
    \caption{Median PLCC / SRCC of compared IQA models in the single-dataset training setting over 10 random splits. All trainable models are trained on KonIQ-10K. \textbf{Bold} indicates best performing model and \underline{underline} indicates next best.}
    \small
    \begin{tabularx}{\textwidth}{l|CCCC|CCC|C}
        \toprule
        \multirow{2}{*}{\textbf{Method}} &
        \multicolumn{4}{c|}{\textbf{Authentic}} &
        \multicolumn{3}{c|}{\textbf{Synthetic}} &
        \multicolumn{1}{c}{\textbf{AIGC}} \\
        \cmidrule(lr){2-5} \cmidrule(lr){6-8} \cmidrule(lr){9-9}
        & \textbf{KonIQ-10K} & \textbf{SPAQ} & \textbf{CLIVE} & \textbf{FLIVE}
        & \textbf{KADID} & \textbf{CSIQ} & \textbf{LIVE} 
        & \textbf{AGIQA-3K} \\
        \midrule
        NIQE \cite{niqe} 
        & 0.533 / 0.530 & 0.679 / 0.664 & 0.493 / 0.449 & 0.147 / 0.100
        & 0.468 / 0.405 & 0.718 / 0.628 & 0.560 / 0.533 & 0.560 / 0.533 \\
        BRISQUE \cite{brisque} 
        & 0.702 / 0.715 & 0.490 / 0.406 & 0.361 / 0.313 & 0.108 / 0.054
        & 0.429 / 0.356 & 0.740 / 0.556 & 0.541 / 0.497 & 0.541 / 0.497 \\
        \midrule
        MANIQA \cite{maniqa}
        & 0.849 / 0.834 & 0.768 / 0.758 & 0.849 / 0.832 & 0.512 / 0.401
        & 0.499 / 0.465 & 0.623 / 0.627 & 0.723 / 0.636 & 0.723 / 0.636 \\
        DBCNN \cite{dbcnn}
        & 0.884 / 0.875 & 0.812 / 0.806 & 0.773 / 0.755 & 0.485 / 0.385
        & 0.497 / 0.484 & 0.586 / 0.572 & 0.730 / 0.641 & 0.730 / 0.641 \\
        NIMA \cite{nima}
        & 0.896 / 0.859 & 0.838 / 0.856 & 0.814 / 0.771 & 0.561 / 0.467
        & 0.532 / 0.535 & 0.695 / 0.649 & 0.715 / 0.654 & 0.715 / 0.654 \\
        HyperIQA \cite{hyperiqa}
        & 0.917 / 0.906 & 0.791 / 0.788 & 0.772 / 0.749 & 0.485 / 0.383
        & 0.506 / 0.468 & 0.752 / 0.717 & 0.702 / 0.640 & 0.702 / 0.640 \\
        CLIP-IQA$^+$ \cite{clipiqa}
        & 0.909 / 0.895 & 0.866 / 0.864 & 0.832 / 0.805 & 0.427 / 0.316
        & 0.653 / 0.654 & 0.772 / 0.719 & 0.736 / 0.685 & 0.736 / 0.685 \\
        MUSIQ \cite{musiq}
        & 0.924 / 0.929 & 0.868 / 0.863 & 0.789 / 0.830 & 0.565 / 0.467
        & 0.575 / 0.556 & 0.771 / 0.710 & 0.722 / 0.630 & 0.722 / 0.630 \\
        \midrule
        C2Score \cite{compare2score}
        & 0.923 / 0.910 & 0.867 / 0.860 & 0.786 / 0.772 & 0.474 / 0.413
        & 0.500 / 0.453 & 0.735 / 0.705 & 0.777 / 0.671 & 0.777 / 0.671 \\
        Q-Align \cite{qalign}
        & 0.941 / 0.940 & 0.886 / 0.887 & 0.853 / 0.860 & 0.554 / 0.483
        & 0.674 / 0.684 & 0.785 / 0.737 & 0.772 / 0.735 & \underline{0.772} / \textbf{0.735} \\
        DeQA \cite{deqa}
        & \underline{0.953} / \underline{0.941} & \textbf{0.895 / 0.896} & \textbf{0.892} / \textbf{0.879} & \underline{0.589} / \underline{0.501}
        & \underline{0.694} / \underline{0.687} & \underline{0.787} / \underline{0.744} & \underline{0.809} / \underline{0.729} & \textbf{0.809} / \underline{0.729} \\
        \midrule
        \textbf{ReLIQS} 
        & \textbf{0.958 / 0.949} & \underline{0.891} / \underline{0.894} & \textbf{0.892} / \underline{0.865} & \textbf{0.654 / 0.549}
        & \textbf{0.701 / 0.707} & \textbf{0.842 / 0.818} & \textbf{0.879 / 0.894} & 0.768 / 0.705 \\
        \bottomrule
    \end{tabularx}
    \label{tab:single_dataset}
\end{table*}
 
\begin{table*}[tb]
    \centering
    \renewcommand{\arraystretch}{0.6}
    \setlength{\tabcolsep}{2.25pt}
    \caption{Median PLCC / SRCC of compared IQA models in the multi-dataset training setting over 10 random splits.}
    \small
    \begin{tabularx}{\textwidth}{l|CCCC|CCC|C}
        \toprule
        \multirow{2}{*}{\textbf{Method}} &
        \multicolumn{4}{c|}{\textbf{Authentic}} &
        \multicolumn{3}{c|}{\textbf{Synthetic}} &
        \multicolumn{1}{c}{\textbf{AIGC}} \\
        \cmidrule(lr){2-5} \cmidrule(lr){6-8} \cmidrule(lr){9-9}
        & \textbf{KonIQ-10K} & \textbf{SPAQ} & \textbf{CLIVE} & \textbf{BID}
        & \textbf{KADID} & \textbf{CSIQ} & \textbf{LIVE} 
        & \textbf{AGIQA-3K} \\
        \midrule
        \multicolumn{9}{l}{\textit{\textbf{Trained on: KonIQ-10K, SPAQ, KADID}}} \\[-2pt]
        \midrule

        Q-Align \cite{qalign}
        & 0.945 / 0.938 & 0.933 / 0.931 & 0.887 / \underline{0.883} & - / - 
        & 0.935 / 0.934 & 0.876 / 0.845 & - / - & 0.788 / \underline{0.733} \\
        DeQA \cite{deqa}
        & \textbf{0.957} / \underline{0.944} & \textbf{0.938} / \textbf{0.934} & \textbf{0.900} / \textbf{0.887} & - / -
        & \textbf{0.955} / \textbf{0.953} & \textbf{0.900} / \textbf{0.857} & - / - & \textbf{0.808} / \textbf{0.745} \\
        \textbf{ReLIQS}
        & \underline{0.954} / \textbf{0.946} & \underline{0.936} / \underline{0.933} & \underline{0.890} / 0.869 & - / -
        & \textbf{0.955} / \textbf{0.953} & \underline{0.894} / \textbf{0.857} & - / - & \underline{0.792} / 0.729 \\

        \midrule
        \multicolumn{9}{l}{\textit{\textbf{Trained on: KonIQ-10K, CLIVE, BID, KADID, CSIQ, LIVE}}} \\[-2pt]
        \midrule
        
        UNIQUE \cite{unique}
        & 0.900 / 0.895 & - / - & 0.884 / 0.854 & 0.875 / 0.852
        & 0.885 / 0.884 & 0.921 / 0.902 & 0.952 / 0.961 & - / - \\
        LIQE \cite{liqe}
        & 0.908 / 0.919 & - / - & 0.910 / 0.904 & 0.900 / 0.875 
        & 0.931 / 0.930 & 0.939 / 0.936 & 0.951 / 0.970 & - / - \\
        Q-Align \cite{qalign}
        & 0.934 / \underline{0.935} & - / - & 0.921 / \textbf{0.931} & 0.920 / 0.904
        & 0.927 / 0.869 & 0.936 / 0.915 & 0.919 / 0.913 & - / - \\
        C2Score \cite{compare2score}
        & \underline{0.939} / 0.931 & - / - & \underline{0.928} / 0.914 & \textbf{0.939} / \textbf{0.919}
        & \underline{0.939} / \textbf{0.952} & \underline{0.943} / \textbf{0.950} & \underline{0.969} / \underline{0.972} & - / - \\
        \textbf{ReLIQS} 
        & \textbf{0.955} / \textbf{0.944} & - / - & \textbf{0.938} / \underline{0.921} & \underline{0.937} / \underline{0.914}
        & \textbf{0.945} / \underline{0.948} & \textbf{0.953} / \underline{0.944} & \textbf{0.980} / \textbf{0.978} & - / - \\
        
        \bottomrule
    \end{tabularx}
    \label{tab:multi-dataset}
\end{table*}

\begin{table}[tb]
    \centering
    \renewcommand{\arraystretch}{0.70}
    \setlength{\tabcolsep}{2.5pt}
    \caption{PLCC / SRCC of compared IQA models on UHD.}
    \small
    \begin{tabularx}{0.8\columnwidth}{l|C|C}
        \toprule
        \textbf{Model} & \textbf{UHD} & \textbf{GMACs}\\
        \midrule
        HyperIQA \cite{hyperiqa}      & 0.103 / 0.553 & 211 \\
        CONTRIQUE \cite{contrique}    & 0.678 / 0.732 & 855 \\
        $\text{CLIP-IQA}^{+}$ \cite{clipiqa} & 0.709 / 0.747 & 895 \\
        GS-PIQA \cite{aim_uhd}        & 0.793 / 0.830 & 50 \\
        SJTU \cite{sjtu}              & 0.799 / 0.846 & 44 \\
        \midrule
        Q-Align \cite{qalign}        & 0.627 / 0.683 & 936 \\
        DeQA \cite{deqa}              & 0.654 / 0.701  & 936 \\
        \midrule
        \textbf{ReLIQS$^*$} & \underline{0.824} / \underline{0.847} & 47 \\
        \textbf{ReLIQS} & \textbf{0.837} / \textbf{0.865} & 543 \\
        \bottomrule
    \end{tabularx}
    \vspace{-8pt}
    \label{tab:uhd_results}
\end{table} 
\subsection{Datasets and Implementation Details}
\paragraph{Datasets.} For a comprehensive evaluation, we trained and evaluated our method on a diverse set of benchmarks covering authentic, synthetic, and AI-generated distortions. Authentically distorted datasets included KonIQ-10K \cite{koniq}, SPAQ \cite{spaq}, CLIVE \cite{clive}, FLIVE \cite{paq2piq}, and BID \cite{bid}; synthetically distorted datasets included KADID \cite{kadid}, LIVE \cite{live}, and CSIQ \cite{csiq}; and the AI-generated dataset was AGIQA-3K \cite{agiqa-3k}. We also trained and evaluated on the UHD \cite{uhd} dataset to assess performance on high-resolution images.

Other than UHD and AGIQA-3K, all datasets were randomly divided into 70/20/10 training, testing, and validation splits. On UHD and AGIQA-3K, we followed the official splits provided by the authors.
\vspace{-8pt}
\paragraph{Implementation Details.} For the patch encoder $E(\cdot)$ in \cref{eq:encoding}, we deployed the CLIP ViT-B/16 with OpenAI weights \cite{clip}. For the IQA-specific importance field network $S(\cdot)$ in \cref{eq:pie-small}, we used the lightweight TinyCLIP ViT-8M \cite{tiny_clip} followed by a shallow convolutional decoder and a softmax across all pixels. 

We trained our model end-to-end using the AdamW optimizer \cite{adamw} and an oscillating cosine learning rate schedule over a period of 10 epochs, an initial and maximum learning rate of $1\times10^{-5}$, and a weight decay of $1\times10^{-3}$. 
The oscillating schedule improves generalization by preventing convergence to sharp minima, particularly in multi-dataset training \cite{sharp_minima,cyclical,sgdr}. We set the margin $\delta$ in \cref{eq:margin-ranking} to be 0.01. We used batch sizes of 16 for KonIQ-10K, SPAQ, and KADID; 4 for BID, CLIVE, CSIQ, and LIVE; and 10 for UHD in both single- and multi-dataset training.
     
We sampled patches from three scales: the original image resolution and two resized variants with short dimensions of 512 and 224 pixels ($R=2$). During training, 6, 5, and 1 patches were randomly sampled from the respective scales, while during evaluation, patches were uniformly sampled with 50\% overlap. We used 4 learned latent quality axes, setting $A=4$.

Following prior work in IQA, we report Spearman’s rank correlation coefficient (SRCC) and Pearson’s linear correlation coefficient (PLCC) as our evaluation metrics.

\subsection{Main Results} 
\label{sec:experiments-main}
In Table \ref{tab:single_dataset}, we compared ReLIQS against recent SOTA NR IQA models. All trainable methods were trained only on KonIQ-10K. Our approach consistently surpassed that of strong deep learning-based baselines such as CLIP-IQA$^+$ and HyperIQA \cite{hyperiqa} across all evaluation benchmarks. Moreover, it also outperformed the transformer-based MUSIQ \cite{musiq}, which, like ReLIQS, fully preserves low-level quality cues with original-resolution processing. 

Notably, our model achieves performance on par with, and in most cases superior to, recent MLLM-based IQA methods including Q-Align \cite{qalign}, Compare2Score \cite{compare2score}, and DeQA \cite{deqa}, while requiring lower computational resources. However, on AGIQA-3K \cite{agiqa-3k}, ReLIQS trails Q-Align and DeQA, possibly due to an AIGC-specific distribution shift under-represented in CLIP pretraining, while MLLM baselines may have greater exposure to generated images.

Table \ref{tab:multi-dataset} reveals our model’s generalization ability in a multi-dataset training setting, showing that it can effectively learn from scarce and heterogeneous IQA labels. When jointly trained on KonIQ-10K, CLIVE, BID, KADID, CSIQ, and LIVE, our method consistently outperformed UNIQUE \cite{unique} and the CLIP-based LIQE \cite{liqe} across all test benchmarks, and exceeded the MLLM-based Compare2Score \cite{compare2score} on nearly all datasets. When trained on KonIQ-10K, SPAQ, and KADID, our model performed on par with or better than Q-Align and DeQA.

Overall, our method achieves strong, mostly SOTA performance in the single-dataset setting and generalizes competitively across diverse content and distortion types, highlighting its effectiveness.

\subsection{High Resolution IQA on UHD Dataset}

Most existing NR-IQA benchmarks, including those in \cref{sec:experiments-main}, are relatively limited in terms of range of resolutions considered. Consequently, globally resizing images to a fixed encoder input usually discards very little task-relevant detail and is thus common practice. The UHD dataset is different: reducing ultra–high-resolution images to typical encoder inputs (e.g., short image dimension 224–448) removes fine structures tied to sharpness, noise, and texture.

Our model retains these cues by sampling patches at original image resolution alongside resized variants, avoiding the information loss of global resizing. We therefore evaluated it on UHD \cite{uhd}, a high-resolution benchmark with authentic distortions, to quantify the benefit of original-resolution patch sampling.

In Table \ref{tab:uhd_results}, we compare ReLIQS with Q-Align and DeQA, all trained on the combined UHD, KonIQ-10K, SPAQ, and KADID datasets. We also included the top two entries from the AIM 2024 Challenge on UHD Blind Photo Quality Assessment \cite{aim_uhd}, namely SJTU \cite{sjtu} and GS-PIQA \cite{aim_uhd}. Although Q-Align and DeQA achieve SOTA results on low- and mid-resolution benchmarks, they perform poorly on UHD, likely because they resize each image to a short dimension of 448 pixels, which suppresses fine details and local structures that are critical for discriminating subtle perceptual differences in UHD images.

The CNN-based baselines CONTRIQUE and CLIP-IQA$^{+}$ do not involve resizing, yet they still exhibit subpar performance on UHD. We attribute this to their ResNet-50 backbones \cite{resnet}, whose limited effective receptive fields make it difficult to capture global semantic structure when the input image is high resolution.

\begin{figure}[t]
  \centering
  \includegraphics[width=0.90\linewidth]{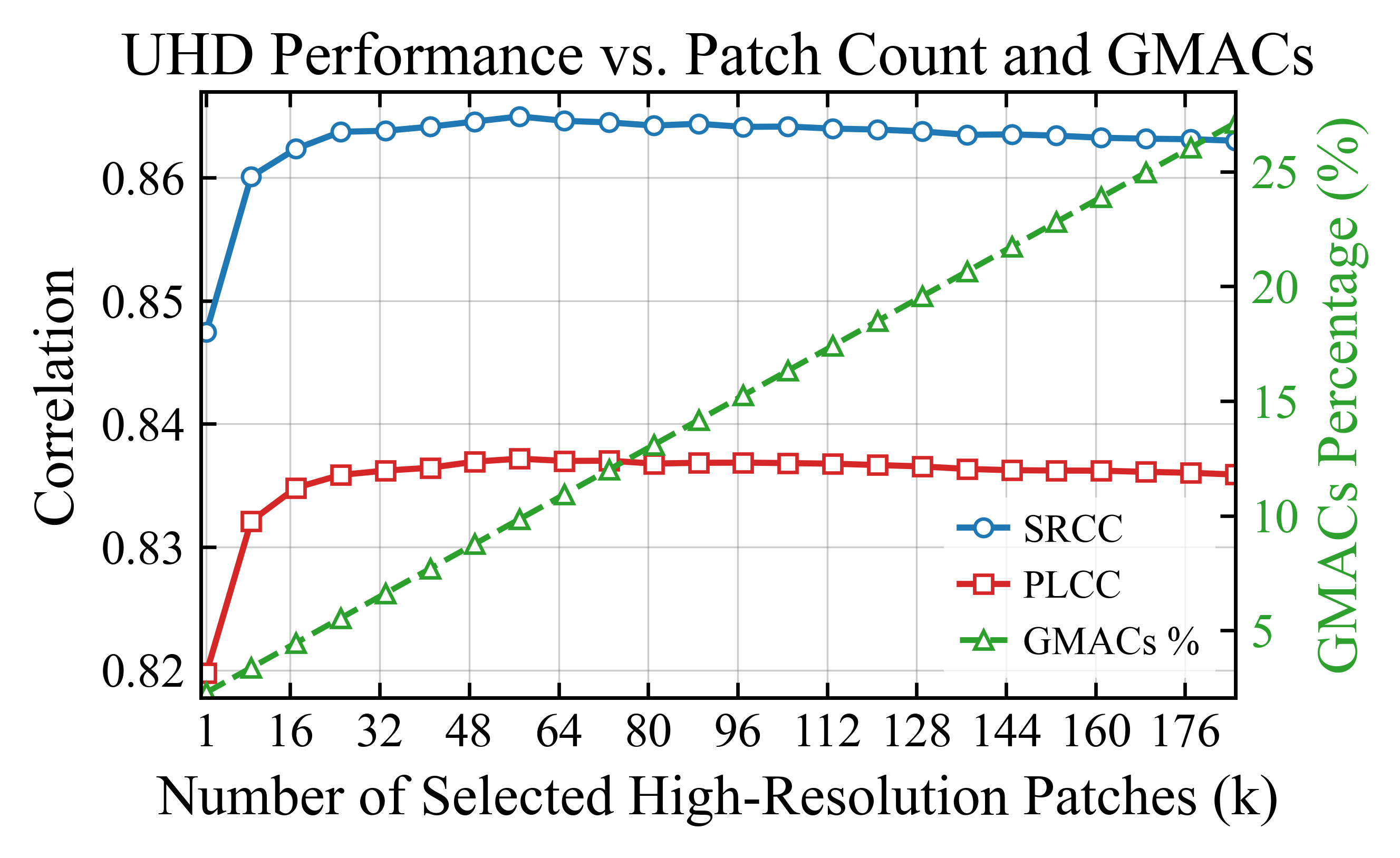}
  \caption{
    PLCC and SRCC with GMACs percentage on UHD with increasing number of selected patches ($k$). Performance saturates rapidly, showing the efficiency of importance-based sampling. 
  }
  \vspace{-6pt}
  \label{fig:srcc_plcc_vs_k}
\end{figure}

ReLIQS, however, captures both low- and high-level quality cues through its multiscale patch sampling and outperforms all compared models in Table \ref{tab:uhd_results} by a significant margin, establishing a new SOTA on UHD. We further introduce a compute-adaptive variant, ReLIQS$^*$, which reduces the cost to 47 GMACs by limiting the number of sampled patches to 4 instead of 48 while outperforming dedicated UHD models GS-PIQA and SJTU, which are specifically tailored to ultra-high-resolution content rather than general-purpose IQA. Our strategy for reducing ReLIQS’s computational cost is detailed in \cref{sec:exp_compute_adaptive}.

\subsection{Compute-Adaptive Patch Selection} 
\label{sec:exp_compute_adaptive} 
\begin{figure}[t]
  \centering
  \includegraphics[width=0.80\linewidth]{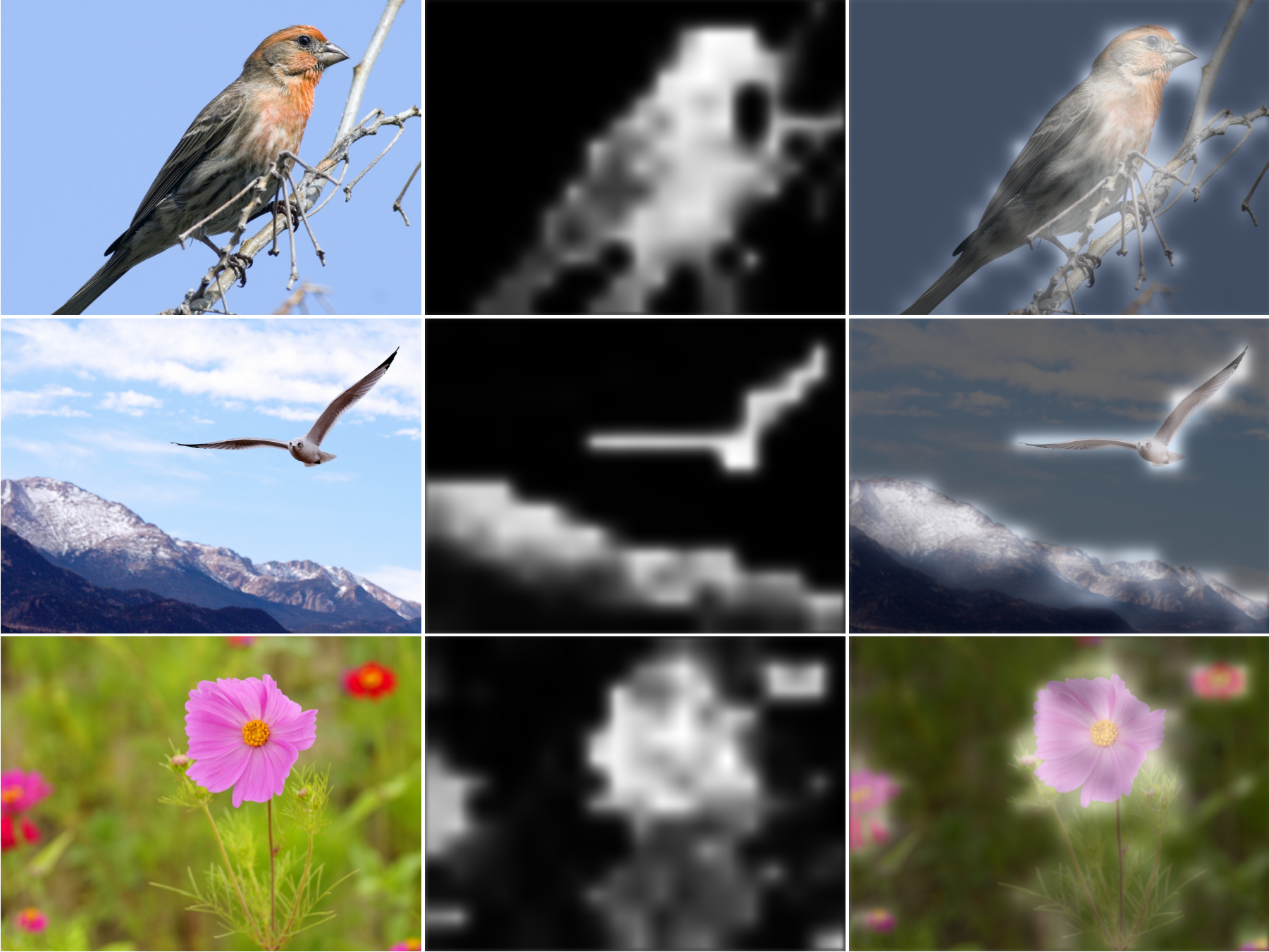}
  \caption{
    Learned IQA-specific saliency maps (middle) with input images (left) and overlays (right). Fine-tuning purely on MOS supervision yields strong bias toward semantically salient regions.
  }
  \vspace{-4pt}
  \label{fig:saliency_comparison}

\end{figure}
 
\begin{table*}[tb]
    \centering
    \renewcommand{\arraystretch}{0.7}
    \setlength{\tabcolsep}{2.25pt}
    \caption{Single-dataset training ablation study on the effects of patch weighting and multi-scale input. KonIQ-10K is used for training. Results are reported as PLCC / SRCC.}
    \small
    \begin{tabularx}{\textwidth}{l|CCCC|CCC|C}
        \toprule
        \multirow{2}{*}{\textbf{Setting}} &
        \multicolumn{4}{c|}{\textbf{Authentic}} &
        \multicolumn{3}{c|}{\textbf{Synthetic}} &
        \multicolumn{1}{c}{\textbf{AIGC}} \\
        \cmidrule(lr){2-5} \cmidrule(lr){6-8} \cmidrule(lr){9-9}
        & \textbf{KonIQ-10K} & \textbf{SPAQ} & \textbf{CLIVE} & \textbf{FLIVE}
        & \textbf{KADID} & \textbf{CSIQ} & \textbf{LIVE} 
        & \textbf{AGIQA-3K} \\
        \midrule

        \multicolumn{9}{l}{\textbf{Patch Weighting}} \\[-2pt]
        \midrule
        Averaging
        & 0.951 / 0.941 & 0.888 / 0.890 & 0.889 / 0.859 & 0.644 / 0.544
        & 0.686 / 0.682 & 0.838 / 0.813 & 0.877 / 0.886 & 0.758 / 0.694 \\
        PIE
        & \textbf{0.958 / 0.949} & \textbf{0.891 / 0.894} & \textbf{0.892 / 0.865} & \textbf{0.654 / 0.549}
        & \textbf{0.701 / 0.707} & \textbf{0.842 / 0.818 } & \textbf{0.879 / 0.894} & \textbf{0.768 / 0.705} \\

        \midrule
        \multicolumn{9}{l}{\textbf{Scales}} \\[-2pt]
        \midrule    
        224
        & 0.904 / 0.871 & 0.882 / 0.891 & 0.874 / 0.826 & 0.648 / 0.538
        & 0.700 / 0.704 & 0.840 / 0.806 & 0.837 / 0.855 & 0.739 / 0.690 \\
        224, 512
        & 0.951 / 0.938 & 0.891 / 0.894 & 0.892 / 0.865 & 0.650 / 0.541
        & 0.701 / 0.707 & 0.842 / 0.818 & 0.879 / 0.894 & 0.768 / 0.705 \\
        224, 512, o.
        & \textbf{0.958 / 0.949} & \textbf{0.891 / 0.894} & \textbf{0.892 / 0.865} & \textbf{0.654 / 0.549}
        & \textbf{0.701 / 0.707} & \textbf{0.842 / 0.818} & \textbf{0.879 / 0.894} & \textbf{0.768 / 0.705} \\

        \bottomrule
    \end{tabularx}
    \label{tab:ablation}
\end{table*}
 
\begin{table}[tb]
    \centering
    \renewcommand{\arraystretch}{0.5}
    \setlength{\tabcolsep}{2.5pt}
    \caption{Multi-dataset training ablation study on the effects of patch weighting, scales, and quality axes on the UHD dataset. UHD, KonIQ-10K, SPAQ, and KADID were used for training. Results are reported as PLCC / SRCC}
    \small
    \begin{tabularx}{0.75\columnwidth}{l|C}
        \toprule
        \textbf{Setting} & \textbf{UHD} \\
        \midrule

        \multicolumn{2}{l}{\textbf{Patch Weighting}} \\[-2pt]
        \midrule
        Averaging & 0.833 / 0.861 \\
        PIE & \textbf{0.837 / 0.865} \\

        \midrule
        \multicolumn{2}{l}{\textbf{Scales}} \\[-2pt]
        \midrule
        224 & 0.686 / 0.680 \\
        224, 512 & 0.756 / 0.750 \\
        224, 512, o. & \textbf{0.837 / 0.865} \\

        \midrule
        \multicolumn{2}{l}{\textbf{Number of Axes}} \\[-2pt]
        \midrule
        1 & 0.828 / 0.858 \\
        2 & 0.834 / 0.859 \\
        4 & \textbf{0.837 / 0.865} \\
        8 & 0.835 / 0.863 \\

        \bottomrule
    \end{tabularx}
    \vspace{-8pt}
    \label{tab:uhd_ablation}
\end{table}
 
On high-resolution images, spatially exhaustive patch sampling and encoding are computationally prohibitive. In our framework, the PIE module predicts IQA-oriented dense importance fields (\cref{eq:pie-small,eq:pie-upsample}) and corresponding normalized patch weights $w_p^{(r)}$ (\cref{eq:saliency_patches,eq:patch_weights}), which can be used to adaptively constrain computation. By ranking patches according to their importance scores, we select and encode only the top-$k$ most informative patches, where $k$ is chosen based on the available computational budget.

In \cref{fig:srcc_plcc_vs_k}, we show the SRCC and PLCC performance along with the percentage of the required compute (GMACs) on the UHD dataset as a function of the number of selected patches ($k$). Candidate patches are first sampled uniformly with 50\% overlap across all scales. For the original-resolution scale, where the candidate pool is large, we retain only the top-$k$ patches with the highest weights $w_p^{(0)}$, renormalize their weights, and perform IQA prediction using these patches. Performance on UHD rapidly saturates with far fewer patches than the full candidate set. For example, a $3840\times2560$ image yields 748 candidates in original resolution, yet the best performance is achieved with only 48 patches—6.4\% of the candidate set.

To visualize the underlying mechanism, \cref{fig:saliency_comparison} illustrates the predicted importance field $\mathbf{s}^{(R)}$. These fields consistently emphasize semantically and perceptually salient regions where humans are most sensitive to distortions \cite{topiq}. As they are obtained through MOS-supervised fine-tuning, we interpret them as \textbf{IQA-specific saliency maps}, related to, but distinct from, visual saliency \cite{eye_tracking}. Further qualitative comparisons along with computational analysis on datasets in Table \ref{tab:single_dataset} are provided in the Supplementary Material.

In summary, the PIE module enables compute-adaptive inference by producing IQA-oriented importance maps that guide patch selection. This mechanism allows the model to concentrate on perceptually critical regions, achieving substantial computational savings while maintaining good performance, particularly for high-resolution images.

\subsection{Ablation Studies}

We performed ablations on the PIE module, patch sampling resolutions, and the number of latent axes to quantify the contribution of each component to overall performance. In Table \ref{tab:ablation}, we reported ablations in the single-dataset setting, using KonIQ-10K for training. Similar to prior work on saliency in IQA \cite{eye_tracking}, we observed a minor but consistent performance gain when mean pooling is replaced by learned IQA-specific saliency (PIE). For patch sampling resolutions, using only the scale with short image dimension 224 leads to a clear performance drop on KonIQ-10K (original resolution $768 \times 1024$), while adding the 512 and original-resolution scales progressively improved performance. On the remaining lower-resolution datasets, additional scales also help, but the gains are smaller, consistent with our earlier argument that resizing is more forgiving for low- and mid-resolution images.        

To better understand the roles of multi-scale sampling and latent quality axes, we performed an additional ablation on UHD in a multi-dataset training setup, jointly training on UHD, KonIQ-10K, SPAQ, and KADID. The results are summarized in Table \ref{tab:uhd_ablation}. On UHD, the impact of adding higher-resolution scales is much stronger than in the single-dataset setting. Using only the scale with short image dimension 224 yielded an SRCC of 0.680, which is 0.185 below the three-scale configuration that includes 224, 512, and the original resolution (0.865 SRCC). Introducing the 512 scale improved SRCC to 0.750, but this still remains 0.115 below the full three-scale model.

Varying the number of latent axes showed a saturating trend. Increasing the number of axes from 1 to 2 and then to 4 consistently improves UHD performance (SRCC 0.858 → 0.859 → 0.865), while using 8 axes offers no further benefit and slightly reduces performance (0.863). On the lower-resolution datasets in Table \ref{tab:ablation}, changing the number of axes produced negligible differences. Taken together, these results suggest that high-resolution content benefits both from multi-scale sampling that preserves native-resolution patches and from modeling a small set of learned quality axes.

\section{Conclusion}
IQA is a critical problem relevant in numerous applications. Despite many SOTA deep learning–based models, it remains challenging to deploy them off-the-shelf in real-world settings. We articulated a set of practical requirements that such models should satisfy: \textbf{(i)} being resolution-agnostic, \textbf{(ii)} preserving low-level quality cues, \textbf{(iii)} having reasonable and budget-adaptive computational cost, and \textbf{(iv)} utilizing supervision from multiple subjective IQA datasets. We proposed a multiscale, CLIP-based framework that is constructed to meet these requirements. In particular, our concept of learned IQA-specific saliency allows us to intelligently limit the number of sampled patches, reducing computational cost with minimal performance drop and addressing \textbf{(iii)}. Extensive experiments on both standard and UHD benchmarks demonstrated that ReLIQS satisfies these criteria in practice and achieves strong performance, while remaining general-purpose. We believe that ReLIQS and its design principles can serve as a reference point for future real-world–directed, general-purpose IQA models that aim to meet these practical requirements.


{
    \small
    \bibliographystyle{ieeenat_fullname}
    \bibliography{main}
}

\clearpage
\setcounter{page}{1}
\maketitlesupplementary

This supplementary material provides: (i) additional details on computational cost and compute-adaptive patch selection; (ii) architectural and implementation details for the IQA-specific saliency module $S(\cdot)$; (iii) additional qualitative saliency visualizations and comparisons with UNISAL; (iv) further experiments on AGIQA-3K; and (v) an ablation study on the training objective.

\subsection{Additional Details on Computational Cost and Compute-Adaptive Patch Selection} 

Unlike most deep learning architectures, ReLIQS does not have a fixed computational cost, since it can operate on a variable number of patches. The only fixed-cost component is the PIE module that produces the perceptual importance map. PIE is intentionally lightweight and operates on a resized version of the input whose short side is 224 pixels, yielding a constant computational cost independent of the original image resolution.

The dominant cost in ReLIQS comes from the patch encoder, and thus scales primarily with the number of selected patches, denoted by $k$ in Sec. 4.4. For low- and mid-resolution images, simple uniform spatial sampling with no overlap or with 50\% overlap remains computationally manageable. For high-resolution images, however, this strategy quickly becomes prohibitive, motivating the need for compute-adaptive patch selection.

In our main experiments on the datasets in Tab. 1, we sample a single patch from the scale with short side 224, and uniformly sample patches with 50\% overlap from the scales with short side 512 and at the original resolution. If the original image resolution is lower than 512, the original-resolution scale and the 512-scale coincide; in that case, we keep only the original-resolution scale, since ReLIQS can operate with an arbitrary number of scales.

\begin{figure}[t]
  \centering
  \includegraphics[width=\linewidth]{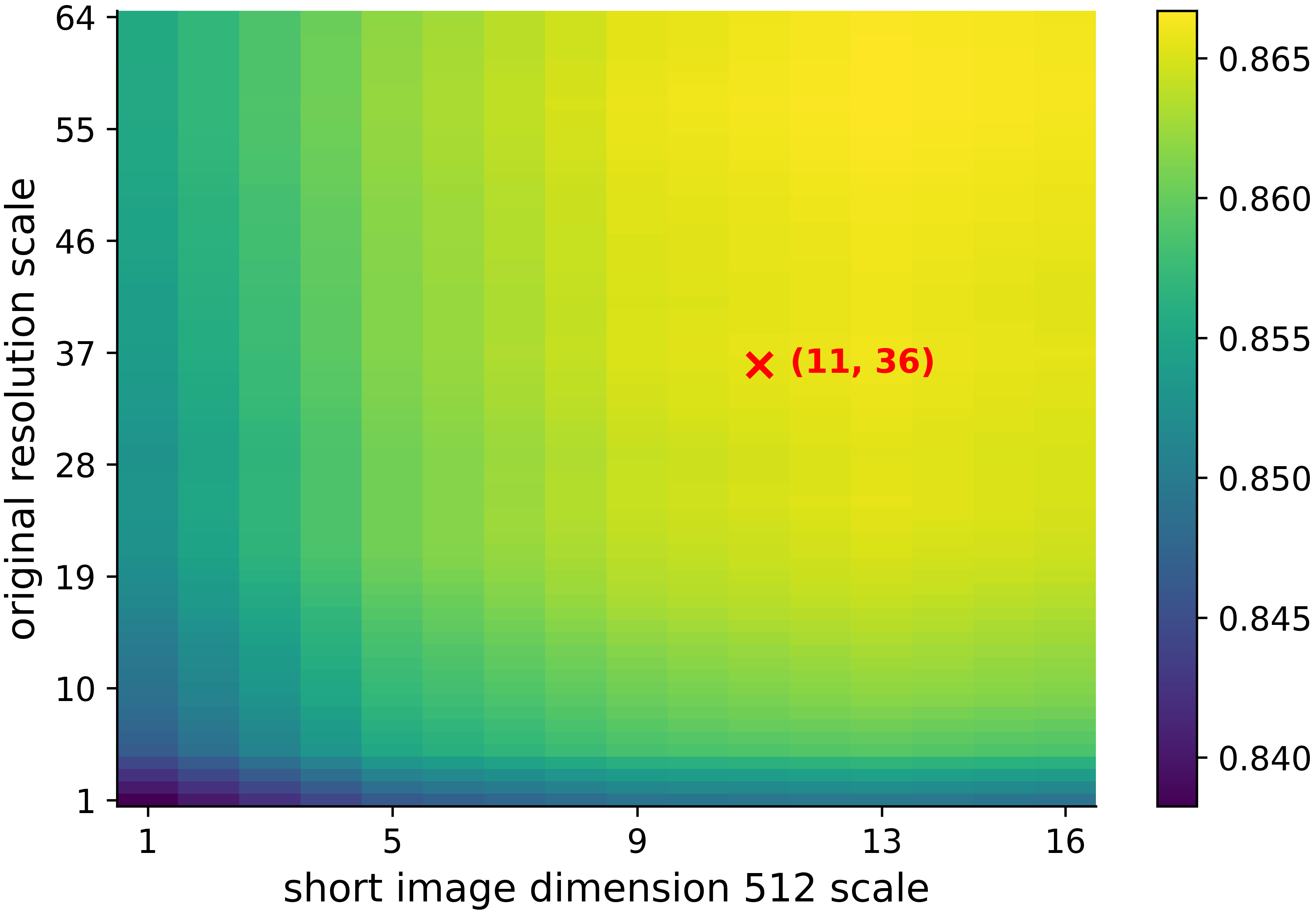}
  \caption{
    SRCC on UHD test set over patch counts at original resolution and short image dimension 512. We select the (11, 36) patch configuration as a good trade-off between performance and computational cost. With one additional patch sampled at short image dimension 224, this corresponds to a total of 48 patches. 
  }
  \label{fig:uhd_landscape}

\end{figure}

\begin{figure}[t]
  \centering
  \includegraphics[width=\linewidth]{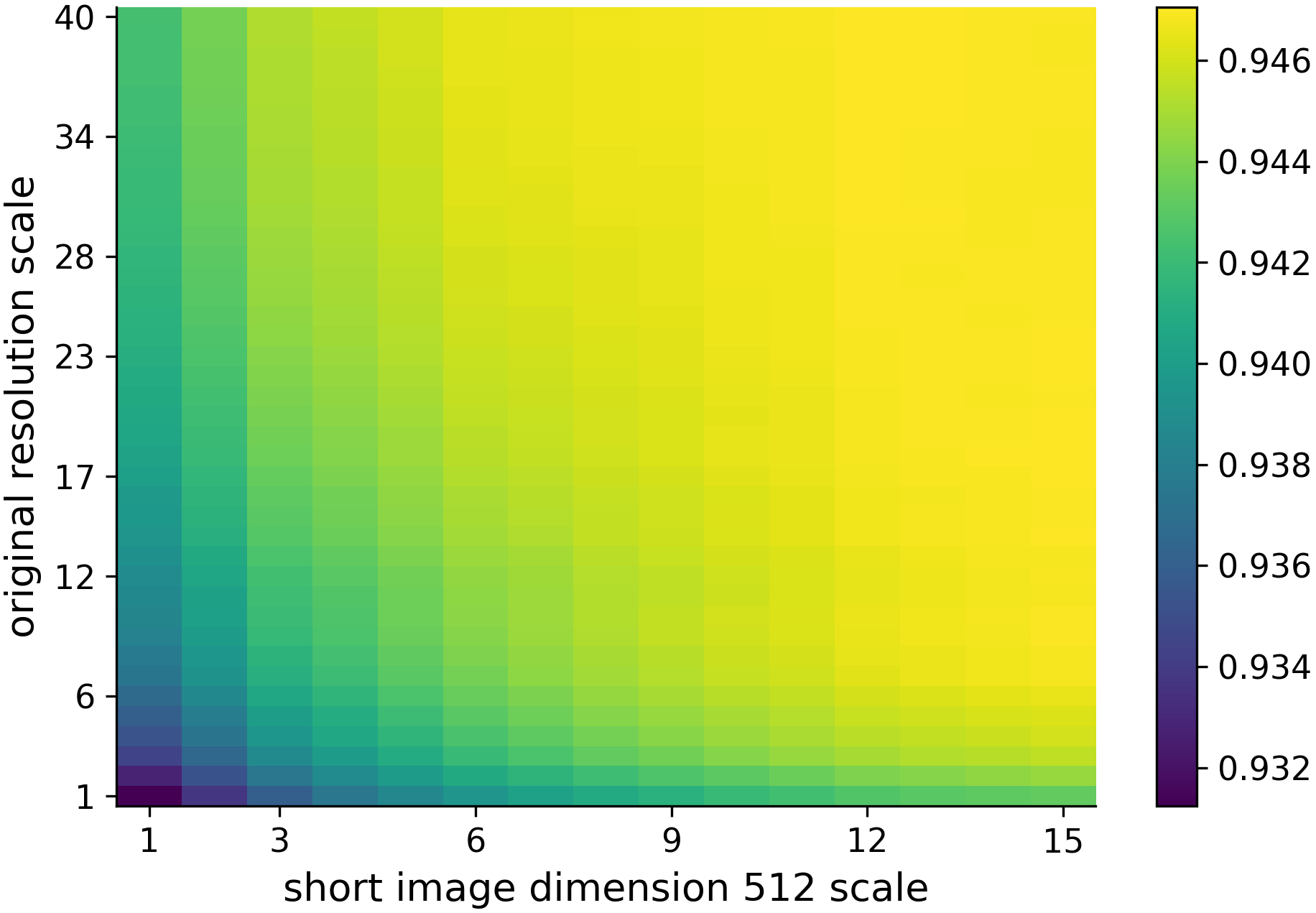}
  \caption{
    SRCC on KonIQ-10K test set over patch counts at original resolution and short image dimension 512. Results are shown for the first split (max SRCC $=0.947$) rather than the median over 10 splits (SRCC $=0.949$).
  }
  \label{fig:koniq_landscape}

\end{figure}

For the UHD dataset in main text Tab. 3, retaining all patches with 50\% overlap is computationally infeasible. We therefore always sample a single patch at the 224-scale and use \cref{fig:uhd_landscape} to decide how many patches to draw from the higher-resolution scales. This figure reports performance as a function of patch count per scale and shows, similar to Fig. 2 in the main text, that performance quickly saturates as the number of patches increases. Utilizing this saturation, we choose a total of $36 + 11 + 1 = 48$ patches per image, allocated from higher to lower scales, respectively, which attains near-maximum performance at substantially reduced computational cost. KonIQ-10K exhibits a similar trend, as shown in \cref{fig:koniq_landscape}. 
 
\begin{table*}[tb]
    \centering
    \renewcommand{\arraystretch}{0.7}
    \setlength{\tabcolsep}{2.25pt}
    \caption{Single-dataset training ablation study on the training objective (PLCC / SRCC). Median performance over 10 splits is reported.}
    \small
    \begin{tabularx}{\textwidth}{l|CCCC|CCC|C}
        \toprule
        \multirow{2}{*}{\textbf{Setting}} &
        \multicolumn{4}{c|}{\textbf{Authentic}} &
        \multicolumn{3}{c|}{\textbf{Synthetic}} &
        \multicolumn{1}{c}{\textbf{AIGC}} \\
        \cmidrule(lr){2-5} \cmidrule(lr){6-8} \cmidrule(lr){9-9}
        & \textbf{KonIQ-10K} & \textbf{SPAQ} & \textbf{CLIVE} & \textbf{FLIVE}
        & \textbf{KADID} & \textbf{CSIQ} & \textbf{LIVE} 
        & \textbf{AGIQA-3K} \\
        \midrule
        \multicolumn{9}{l}{\textit{\textbf{Trained on: KonIQ-10K}}} \\[-2pt]
        \midrule
        Only PLCC 
        & \textbf{0.958} / 0.941 & \textbf{0.891} / 0.883 & 0.891 / 0.856 & 0.653 / 0.540 
        & 0.700 / 0.691 & \textbf{0.842} / 0.810 & \textbf{0.879} / 0.880 & \textbf{0.770} / 0.695 \\
        Only MR 
        & 0.924 / 0.947 & 0.878 / 0.888 & 0.850 / \textbf{0.865} & 0.601 / 0.548
        & 0.695 / 0.705 & 0.794 / 0.813 & 0.860 / 0.871 & 0.752 / \textbf{0.705} \\
        PLCC + MR (uncer.) 
        & \textbf{0.958 / 0.949} & \textbf{0.891 / 0.894} & \textbf{0.892 / 0.865} & \textbf{0.654 / 0.549}
        & \textbf{0.701 / 0.707} & \textbf{0.842 / 0.818 } & \textbf{0.879 / 0.894} & 0.768 / \textbf{0.705} \\
        \midrule
        \multicolumn{9}{l}{\textit{\textbf{Trained on: KonIQ-10K, SPAQ, KADID}}} \\[-2pt]
        \midrule 
        PLCC + MR (equal) & 0.953 / 0.944 & 0.932 / 0.928 & \textbf{0.890} / 0.867 & - / - & 0.952 / 0.950 & 0.893 / 0.855 & - / - & 0.758 / 0.714\\
        PLCC + MR (uncer.) & \textbf{0.954} / \textbf{0.946} & \textbf{0.936} / \textbf{0.933} & \textbf{0.890} / \textbf{0.869} & - / -
        & \textbf{0.955} / \textbf{0.953} & \textbf{0.894} / \textbf{0.857} & - / - & \textbf{0.792} / \textbf{0.729} \\ 
        \bottomrule
    \end{tabularx}
    \label{tab:supp_loss_ablation}
\end{table*}
 
\begin{table}[tb]
    \centering
    \renewcommand{\arraystretch}{0.7}
    \setlength{\tabcolsep}{2.5pt}
    \caption{AGIQA-3K fine-tuning results on the test set (PLCC / SRCC).}
    \small
    \begin{tabularx}{0.8\columnwidth}{l|C}
        \toprule
        \textbf{Model} & \textbf{AGIQA-3K} \\
        \midrule
        OneAlign + LoRA \cite{qalign}  & 0.920 / 0.880 \\
        \textbf{ReLIQS} & \textbf{0.933} / \textbf{0.892} \\
        \bottomrule
    \end{tabularx}
    \vspace{-8pt}
    \label{tab:supp_agiqa}
\end{table} 

Although we report our main results in main text Tabs. 1 and 2 using the full 50\% overlap patch set for simplicity, the corresponding curves indicate that the number of patches can be reduced substantially for KonIQ images with resolution $768 \times 1024$ while maintaining performance close to the reported scores. Overall, these analyses demonstrate that ReLIQS can flexibly trade computational cost for accuracy by adjusting the patch budget per scale.

Unless otherwise noted, we compute GMACs using the \texttt{thop} library for both ReLIQS and all baseline models.

\subsection{Details on IQA-Specific Saliency} 

The lightweight network $S(\cdot)$ produces perceptual importance maps, which we refer to as IQA-specific saliency maps, from which relative patch weights are derived. $S(\cdot)$ uses a TinyCLIP ViT-8M visual encoder followed by a shallow convolutional head with two depthwise $5 \times 5$ convolutions, each followed by a GeLU nonlinearity, and a final pointwise $1 \times 1$ convolution. Perceptual importance maps are obtained by applying a softmax over all spatial locations to normalize the saliency values before patch sampling.

The input to $S(\cdot)$ is the image resized so that its short side is 224 pixels. Since the long side can vary, we interpolate the learned positional embeddings of the TinyCLIP ViT-8M visual encoder to the corresponding spatial grid before tokenization. When the long side is not divisible by the internal patch size (16 pixels), we pad the image, run the encoder, and then crop the padded regions from the dense feature map before feeding it to the convolutional head. Although both $S(\cdot)$ and $E(\cdot)$ use transformer-based CLIP backbones, $E(\cdot)$ outputs only the CLS token, whereas $S(\cdot)$ utilizes the full set of spatial (dense) features. 

\subsection{Additional Qualitative Results on IQA-Specific Saliency}
$S(\cdot)$ outputs normalized saliency maps at a short side of 224 pixels. In the main text Fig. 3, we visualize these maps to highlight regions the model deems perceptually important. For visualization, we take the output of $S(\cdot)$, scale it by 255, round, cast to \texttt{uint8}, and replicate the single-channel map across the three RGB channels. The resulting visualizations appear as grayscale maps, where brighter regions correspond to higher predicted importance. In \cref{fig:supp_saliency,fig:supp_saliency_2}, we show additional examples of these IQA-specific saliency maps. 

\begin{figure*}[t]
  \centering
  \includegraphics[width=\linewidth]{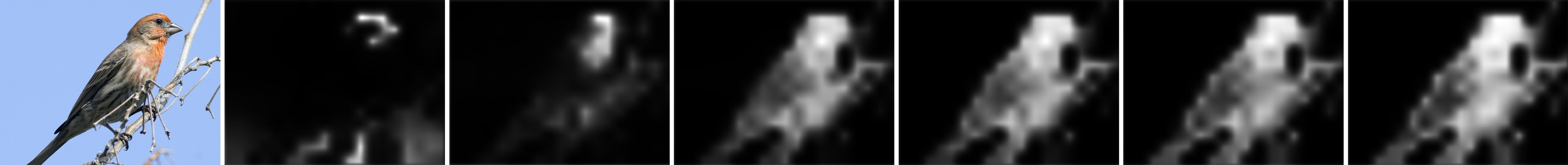}
  \caption{
    Learning progression of IQA-specific saliency maps for models trained on KonIQ-10K, CLIVE, BID, KADID, CSIQ, and LIVE. Columns show the original image, $S(\cdot)$ output before training, and outputs after 10, 20, 30, 40, and 50 epochs (the best checkpoint for this run is at epoch 51). The maps become progressively more structured and semantically aligned over training. 
  }
  \label{fig:saliency_progression}

\end{figure*}

\begin{figure*}[t]
  \centering
  \includegraphics[width=\linewidth]{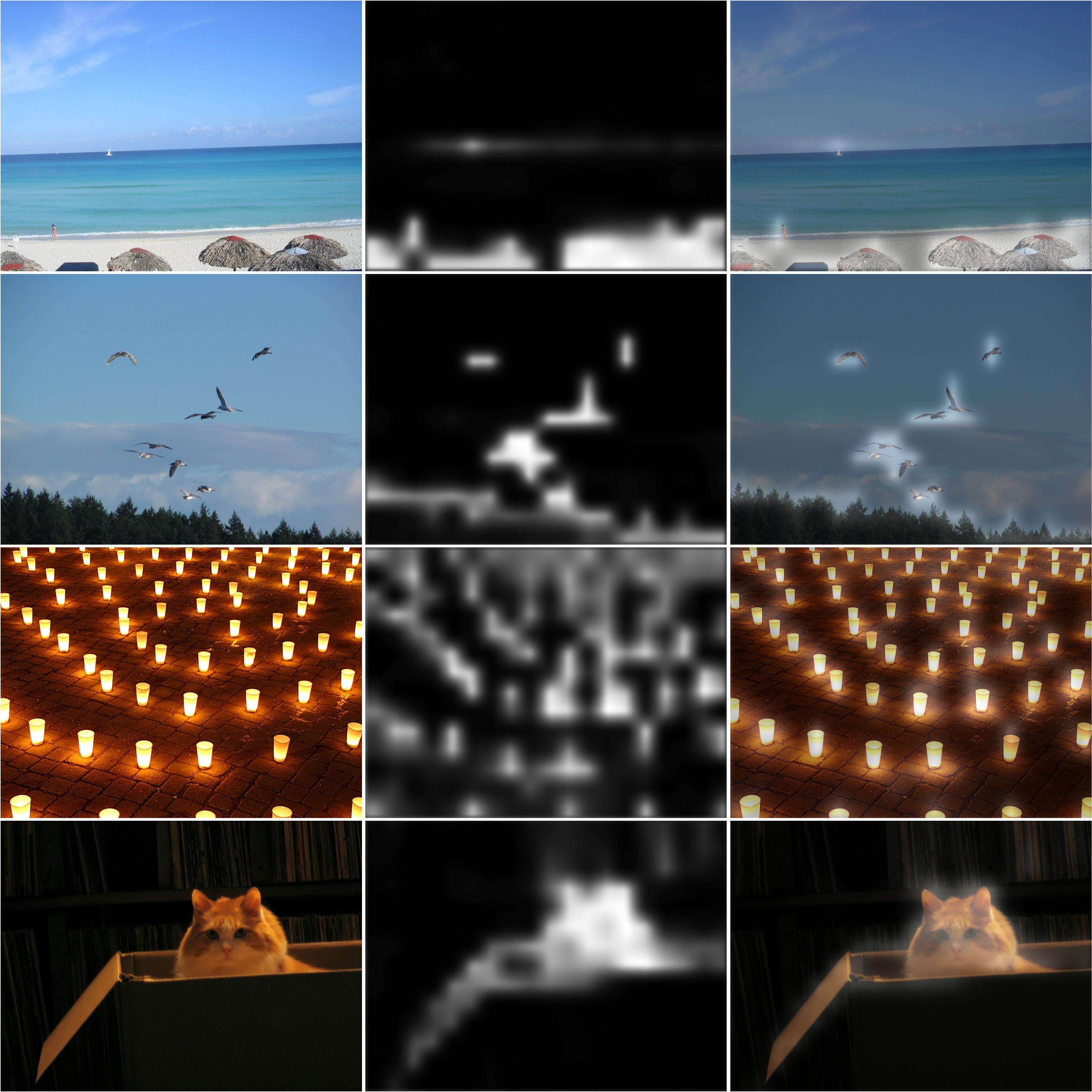}
  \caption{
    Learned IQA-specific saliency maps (middle) with input images (left) and overlays (right). Fine-tuning purely on MOS supervision yields strong bias toward semantically salient regions.
  }
  \label{fig:supp_saliency}

\end{figure*}

\begin{figure*}[t]
  \centering
  \includegraphics[width=\linewidth]{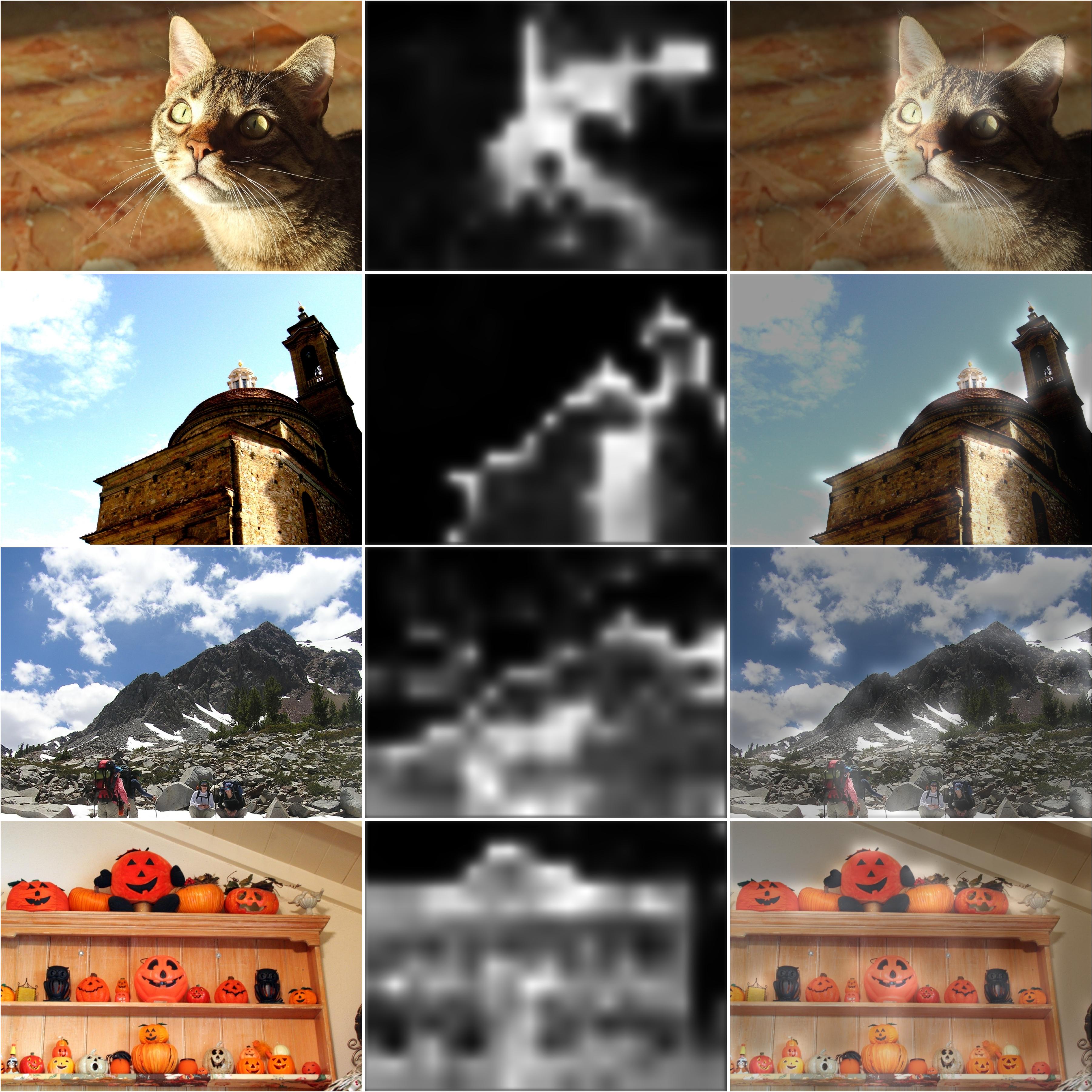}
  \caption{
    Learned IQA-specific saliency maps (middle) with input images (left) and overlays (right). Fine-tuning purely on MOS supervision yields strong bias toward semantically salient regions.
  }
  \label{fig:supp_saliency_2}

\end{figure*}

Alongside additional qualitative results, we visualize the learning progression of the IQA-specific saliency maps. Specifically, we plot $S(\cdot)$ predictions in the multi-dataset setting, where the model is trained on KonIQ-10K, CLIVE, BID, KADID, CSIQ, and LIVE. As shown in \cref{fig:saliency_progression}, the maps evolve from diffuse responses to more structured, semantically aligned patterns as training proceeds.  

\begin{figure*}[t]
  \centering
  \includegraphics[width=\linewidth]{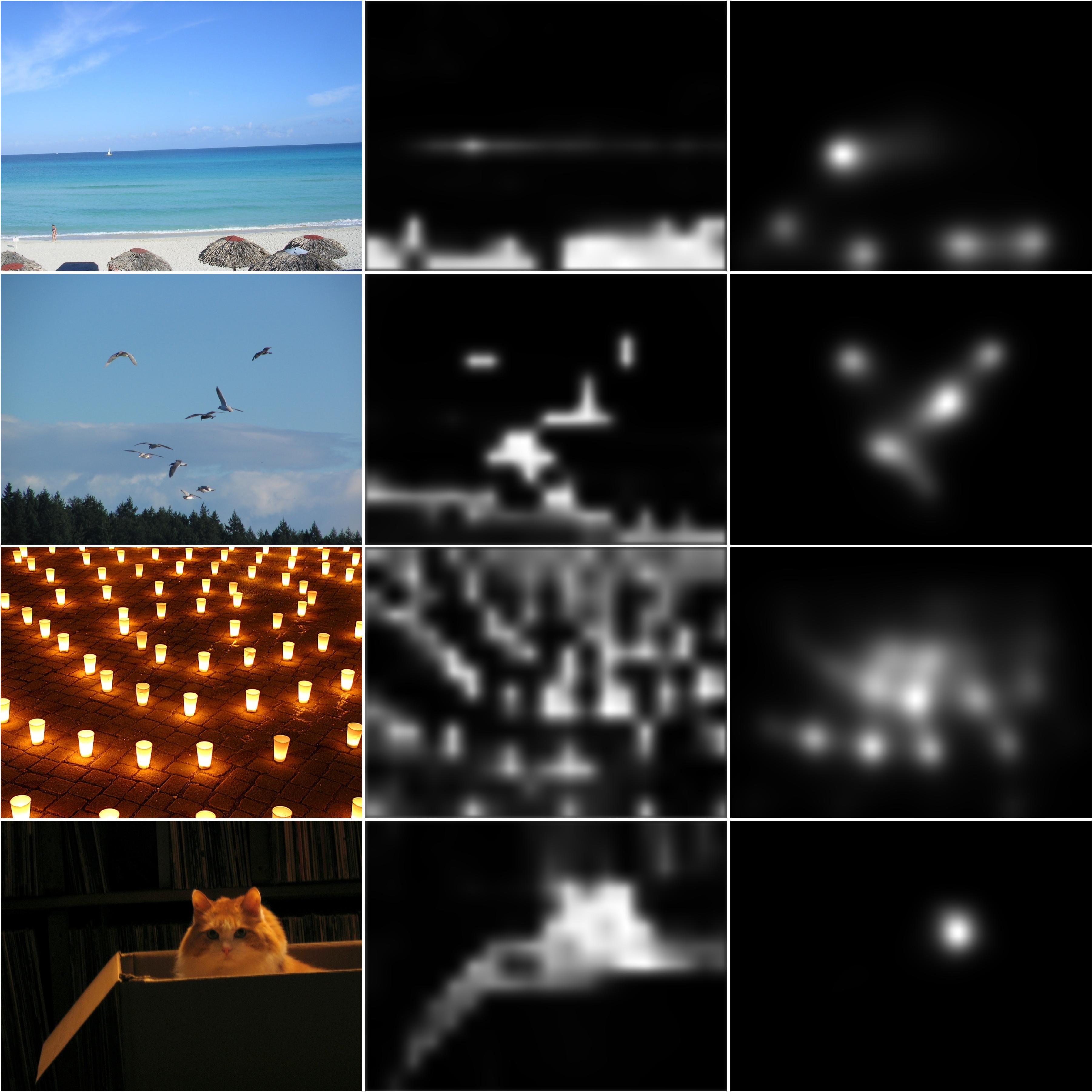}
  \caption{
    Input images (left), learned IQA-specific saliency maps (middle), and UNISAL visual saliency maps (right). In these examples, the IQA-specific saliency maps highlight broader regions, while the UNISAL maps are more spatially concentrated.
  }
  \label{fig:supp_unisal_comparison}

\end{figure*}

We also compared the learned IQA-specific saliency with conventional visual saliency using the UNISAL model \cite{unisal}. In the examples shown in \cref{fig:supp_unisal_comparison}, the UNISAL saliency maps appear sharper and more spatially concentrated, whereas the IQA-specific saliency maps tend to highlight broader regions.

\subsection{Further Evaluation on AGIQA-3K Dataset} 

Recall that in main text Tab. 1, with all models trained only on KonIQ-10K, the cross-dataset performance of ReLIQS on AGIQA-3K lags behind Q-Align and DeQA. In the main paper, we attributed this gap to pretraining differences: ReLIQS uses the OpenAI CLIP ViT-B/16 visual encoder, whereas Q-Align and DeQA build on more recent MLLMs that are more likely to have encountered AI-generated content during pretraining.

To probe this, we trained ReLIQS only on AGIQA-3K and compared it against the reported performance of OneAlign with LoRA fine-tuning on AGIQA-3K, where OneAlign is the Q-Align variant jointly trained on multiple IQA and video quality assessment datasets. Unless otherwise noted, we use the same training settings as in the main experiments (optimizer, learning rate schedule, etc.), but restrict supervision to AGIQA-3K. As shown in \cref{tab:supp_agiqa}, ReLIQS outperforms the OneAlign + LoRA variant by +1.3 PLCC and +1.2 SRCC on the official test split. These results indicate that our architecture is not inherently ill-suited to AI-generated image quality assessment and can be effectively repurposed for this setting via fine-tuning. They are also consistent with our view that the cross-dataset gap in main text Tab. 1 stems, at least in part, from differences in pretraining exposure and training data rather than fundamental architectural limitations.

\subsection{Ablation Study on the Training Objective}
Our training objective combines a margin-ranking term $\mathcal{L}_\text{MR}$ and a PLCC term $\mathcal{L}_\text{PLCC}$, with their relative weights learned via an uncertainty-based scheme. In this scheme, we initialize all loss-specific scales in Eq. (14) by setting the corresponding $\sigma$ values to 1.

In \cref{tab:supp_loss_ablation}, we ablate the training objective in both single-dataset and multi-dataset setups. In the single-dataset setting on KonIQ-10K, combining $\mathcal{L}_\text{PLCC}$ and $\mathcal{L}_\text{MR}$ consistently improves performance over using either term alone. Using only $\mathcal{L}_\text{PLCC}$ yields results closer to the combined case, but with noticeably lower SRCC, whereas using only $\mathcal{L}_\text{MR}$ attains SRCC comparable to the combined setting while substantially degrading PLCC.

In the multi-dataset setting, where the model is trained on KonIQ-10K, SPAQ, and KADID, uncertainty-based adaptive weighting of $\mathcal{L}_\text{PLCC}$ and $\mathcal{L}_\text{MR}$ provides a small but consistent performance gain over a simple variant that weights the two losses equally. In the single-dataset setting, however, this gain was negligible, likely because the model converges earlier when trained on a single dataset and we apply early stopping. Overall, the uncertainty-based combination of loss terms yields a slight but consistent performance boost, particularly in the multi-dataset setting.

\subsection{Further Comparisons with SOTA IQA Methods} 
\begin{table}[tb]
    \centering
    \renewcommand{\arraystretch}{0.75}
    \setlength{\tabcolsep}{5.0pt}
    \caption{Median PLCC / SRCC of compared IQA models in the single-dataset training setting. Models are trained on KonIQ-10K. \textbf{Bold} indicates best performing model and \underline{underline} indicates the next best.}
    \small
    \begin{tabular}{l|c}
        \toprule
        \textbf{Method} & \textbf{KonIQ-10K} \\
        \midrule
        GrepQ \cite{grepq} & - / 0.855 \\
        QPT \cite{qpt} &  0.941 / 0.927 \\
        LODA \cite{loda} & 0.944 / 0.932 \\
        QCN \cite{qcn} &  0.945 / 0.934 \\
        SHDIQA \cite{shdiqa} & 0.948 / 0.937 \\  
        ATTIQA \cite{attiqa} & \underline{0.952} / \underline{0.942} \\  
        \midrule
        \textbf{ReLIQS} & \textbf{0.958 / 0.949} \\
        \bottomrule
    \end{tabular}
    \label{tab:further_comparison}
\end{table} 

In addition to the baselines in the main text, we compared ReLIQS against several recently proposed SOTA IQA methods. When trained and evaluated on KonIQ-10K, as tabulated in \cref{tab:further_comparison}, ReLIQS outperformed ATTIQA \cite{attiqa}, SHDIQA \cite{shdiqa}, QCN \cite{qcn}, and LODA \cite{loda} by PLCC/SRCC margins of 0.006/0.007, 0.010/0.012, 0.013/0.015, and 0.014/0.017, respectively.

\end{document}